\RequirePackage{filecontents}

\documentclass[pdflatex,sn-mathphys]{sn-jnl}

\usepackage{lmodern}
\usepackage{fix-cm}
\usepackage{mathrsfs}
\usepackage{float}

\usepackage{multirow}
\usepackage{comment}
\usepackage{commath}
\usepackage{soul}
\usepackage{bookmark}
\usepackage[capitalize,noabbrev]{cleveref}
\usepackage{textcomp}

\usepackage{pdfpages}
\definecolor{darkred}{RGB}{200,0,0} 
\newcommand{\revision}[1]{%
  \begingroup
    \color{darkred}#1%
  \endgroup%
}
\renewcommand{\revision}[1]{#1}

\jyear{2021}%
\theoremstyle{thmstyleone}%
\theoremstyle{thmstyletwo}%

\theoremstyle{thmstylethree}%

\usepackage{lineno}
\linenumbers

\begin{document}
\nolinenumbers

\title[Article Title]{Riemannian Denoising Model for Molecular Structure Optimization with Chemical Accuracy}

\author[1]{\fnm{Jeheon} \sur{Woo}}\email{jeheon@kisti.re.kr}
\equalcont{These authors contributed equally to this work.}
\author[1,3]{\fnm{Seonghwan} \sur{Kim}}\email{dmdtka00@kaist.ac.kr}
\equalcont{These authors contributed equally to this work.}
\author[1]{\fnm{Jun Hyeong} \sur{Kim}}\email{junhkim1226@kaist.ac.kr}
\author*[1,2,3]{\fnm{Woo Youn} \sur{Kim}}\email{wooyoun@kaist.ac.kr}

\affil[1]{\orgdiv{Department of Chemistry}, \orgname{KAIST}, \orgaddress{\street{291 Daehak-ro}, \city{Yuseong-gu}, \postcode{34141}, \state{Daejeon}, \country{Republic of Korea}}}
\affil[2]{\orgdiv{Graduate School of Data Science}, \orgname{KAIST}, \orgaddress{\street{291 Daehak-ro}, \city{Yuseong-gu}, \postcode{34141}, \state{Daejeon}, \country{Republic of Korea}}}
\affil[3]{\orgdiv{InnoCORE AI-CRED Institute}, \orgname{KAIST}, \orgaddress{\street{291 Daehak-ro}, \city{Yuseong-gu}, \postcode{34141}, \state{Daejeon}, \country{Republic of Korea}}}

\abstract{We introduce a framework for molecular structure optimization using denoising model on a physics-informed Riemannian manifold (R-DM). Unlike conventional approaches operating in Euclidean space, our method leverages a Riemannian metric that better aligns with molecular energy change, enabling more robust modeling of potential energy surfaces. By incorporating internal coordinates reflective of energetic properties, R-DM achieves chemical accuracy with an energy error below 1 kcal/mol. Comparative evaluations on QM9, QM7-X, and GEOM datasets demonstrate improvements in both structural and energetic accuracy, surpassing conventional Euclidean-based denoising models.  This approach highlights the potential of physics-informed coordinates for tackling complex molecular optimization problems, with implications for tasks in computational chemistry and materials science.}
\keywords{Molecular conformation, Structure optimization, Machine learning, Score matching, Diffusion model, Riemannian manifold}

\maketitle
\section{Introduction}\label{sec1}
Accurate prediction of molecular structures is pivotal in various scientific fields, ranging from drug discovery to materials science \cite{jorgensen2004many,jain2013commentary}. Understanding the spatial arrangement of atoms in molecules not only elucidates their properties, but also facilitates the design of novel compounds with the desired characteristics \cite{cramer2013essentials,leach2001molecular}. Traditionally, quantum mechanical simulations, such as density functional theory (DFT), have been the primary approach for predicting molecular structures \cite{jones2015density,YangParr}. However, the inherent complexity and high computational cost of these simulations often limit their practicality \cite{friesner2005ab,bowler2012methods,goedecker1999linear,woo2022system}. These limitations present substantial challenges in extending these methods to larger and more complex compounds, highlighting the need for more efficient and scalable predictive approaches \cite{rupp2012fast,butler2018machine,von2018quantum,woo2023dynamic,woo2024efficient}.

In recent years, machine learning techniques have emerged as powerful tools in the field of molecular modeling with substantial potential to overcome these challenges \cite{smith2017ani,GeoDiff,TSDiff}. Previous research has demonstrated the efficacy of machine learning in predicting molecular properties and structures, opening a new door for innovative applications in diverse science domains. One category of molecular structure prediction involves learning a potential energy surface (PES), where models aim to predict the potential energy and associated forces of a given molecular structure \cite{dral2020hierarchical, chmiela2018towards, smith2017ani, behler2007generalized, rupp2012fast}. These methods rely on auxiliary optimization algorithms, similar to conventional computational chemistry approaches, to find molecular structures by navigating the energy landscape. 

Alternatively, models for predicting molecular structures have focused on learning the data distributions of atomic positions \cite{xu2021end,mansimov2019molecular,ConfGF,DGSM,SDEGen,DMCG,GeoDiff,TSDiff,OA-ReactDiff,simm2019generative}. Variational autoencoders, for example, offer one-shot generation of molecular conformers by predicting internal coordinates such as interatomic distances and angles \cite{xu2021end,mansimov2019molecular}. More recently, denoising score matching (SM) and flow matching (FM) frameworks have been applied to molecular conformer modeling. Both denoising models (DM) learn a vector field to drive a prior noisy molecular structure toward its clean counterpart \cite{ConfGF, DGSM, SDEGen,hassan2024flow}, thus transporting a prior distribution to a target distribution. The denoising process is conducted by integrating differential equations (stochastic or ordinary) and is analogous to conventional molecular structure optimization algorithms, where the vector field acts like physical force fields steering molecular structures toward their minimum energy states. Most existing DMs operate in Euclidean space, where atomic positions are represented in Cartesian coordinates, and denoising models predict direct changes in the Cartesian coordinates. These approaches have shown state-of-the-art performance in various conformer predictions \cite{GeoDiff,TSDiff,OA-ReactDiff,duan2025optimal,hassan2024flow,morered}.

In parallel with advances in models equipped with Euclidean space, recent works have proposed generative models that explicitly incorporate Riemannian manifolds to represent the intrinsic degrees of freedom and roto-translational configuration in molecular structures \cite{jing2022torsional,corso_diffdock_2023,yim2023fast}. These models define DMs on constrained geometric spaces, such as spheres and hypertori, acting on specific coordinates while leaving the other degrees of freedom fixed. This focus on selected degrees of freedom aligns with a broader trend in molecular modeling, where physically meaningful subspaces, such as bond lengths, angles, dihedrals, or collective variables, are isolated to improve interpretability and geometric coherence. Even beyond generative modeling, recent studies have adopted Riemannian geometric frameworks to capture and analyze structural variability within efficiently reduced coordinate subspaces \cite{diepeveen2024riemannian,yang2024learning}, and to guide transition path sampling by incorporating modeled energy landscapes \cite{diepeveen2024riemannian,yang2024learning,wang2024generalized}.

Achieving chemical accuracy, an energy error below 1 kcal/mol, has long been considered a benchmark for data-driven structure prediction. Until very recently, this target remained elusive, but the latest DMs have demonstrated the capability to achieve this target on several standard test sets \cite{morered,hassan2024flow,duan2025optimal}. We pursue the same goal from a complementary perspective: rather than refining model capacity or loss functions, we redefine the geometry of the space so that structural displacement aligns with the energetic change. Most current methods instead rely on Cartesian $L^{2}$ or torsion-angle spaces, in which every direction is treated equally. However, equal metric distances do not correspond to equal energy changes due to the inherent complexity of the energy landscape, and the learned vector field may therefore transport a molecular structure without faithfully minimizing its energy.

\revision{Inspired by the Fukui's intrinsic reaction coordinate (IRC) theory \cite{tachibana1979intrinsic}}, we address this mismatch by introducing a physics-informed Riemannian manifold whose metric is explicitly correlated with the local curvature of the potential energy surface. Building on the physics-informed internal coordinates of Zhu \textit{et al.}\ \cite{zhu2019geodesic}, we warp interatomic distances so that displacements along stiff bonds cost more than those along soft torsions, thereby aligning geodesic distance with the energy gap. Training the DM with a Gaussian perturbation on the manifold allows the denoising vector field to more closely approximate the actual force field near the clean structure, enabling the model to accurately learn the distribution of molecular structures while ensuring that the generated structures are energetically accurate. Through a comparative analysis of the noisy structures from both Euclidean and Riemannian spaces, we confirmed a substantial divergence between them and demonstrated that the designed Riemannian manifold reflects the potential wells of molecular conformations. In the denoising step, since prior distributions on a Riemannian manifold are generally intractable, we adopt a formulation that mimics the traditional optimization scheme; our approach starts with a suboptimal initial guess and then denoises it toward the corresponding energy minimum state. This approach bypasses the need to explicitly define an intractable prior while still capturing the potential wells. As a result, the denoising step effectively functions as a molecular structure optimization procedure.

\revision{To demonstrate the performance of Riemannian DM (R-DM), we first evaluated it on the QM9 dataset \cite{qm9}, where it outperformed the Euclidean counterpart with a median energy error of 0.177 kcal/mol from force-field–initialized structures. We then assessed R-DM on QM7-X \cite{qm7x}, confirming its advantage in relaxing non-equilibrium structures generated by normal-mode sampling. On the GEOM-QM9 subset of the GEOM dataset \cite{GEOM}, R-DM further refined conformations from ETKDG and generative models \cite{ConfGF,GeoDiff}, improving both structure-based metrics and ensemble properties. Together, these results demonstrate the effectiveness of R-DM and its ability to achieve chemical accuracy, a key criterion for reliable molecular modeling.}

\section{Results}\label{sec2}

\subsection{Design of physics-informed Riemannian metric}\label{subsec2-2}
\begin{figure}[!h]
\centering
\includegraphics[width=1.0\textwidth]{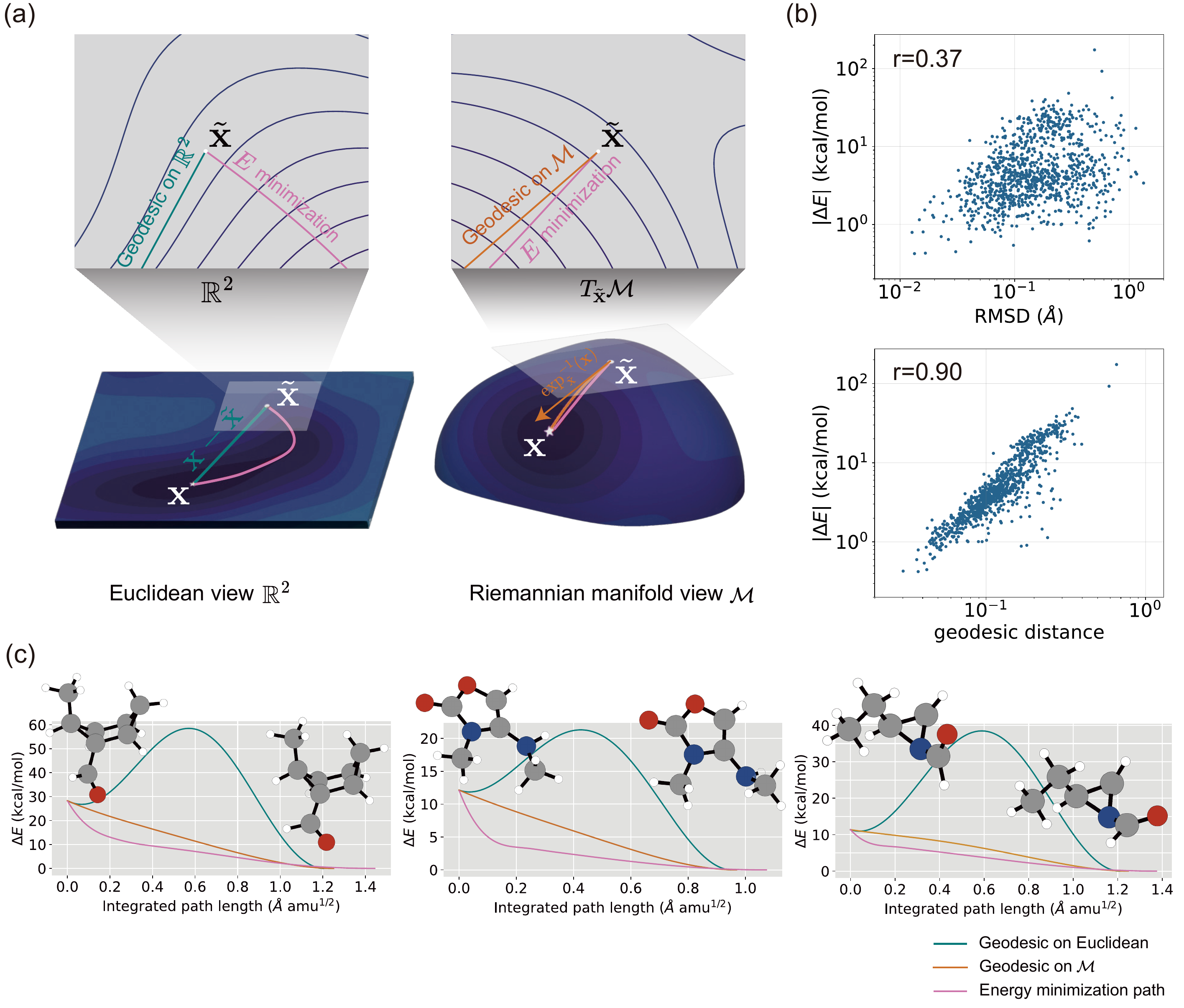}
\caption{
\textbf{Comparative analysis of Euclidean and Riemannian spaces in molecular structure optimization.}
\textbf{a}
Conceptual illustration of the potential energy surface (PES) in Euclidean $\mathbb{R}^2$ and Riemannian spaces $\mathcal{M}$. The schematic highlights the differences between the geodesics in each space (teal and orange lines) and the energy minimization path (magenta lines) on the PES. In the panels below, the direction vectors from $\tilde{\mathbf{x}}$ to $\mathbf{x}$, expressed on the tangent plane, are shown.
\textbf{b}
Correlation of energy differences, $\abs{\Delta E}$, with the two measures of structural differences: root mean square deviation (RMSD) and geodesic distance. The energy and structure differences were calculated for 1,000 randomly selected pairs of structures optimized by the MMFF and DFT methods. Both the x- and y-axes are plotted on a log-log scale and have been adjusted for better visualization. Pearson's correlation coefficients (r) quantify the linear relationship between the energy differences and each structural measurement.
\textbf{c}
Energy profiles are shown with the $x$-axis representing the integrated path length, calculated as the cumulative sum of RMSD values between neighboring structures in mass-weighted Cartesian coordinates. Teal and orange lines denote the geodesics in Euclidean and Riemannian spaces, respectively, while the magenta line indicates the energy minimization path. The left and right endpoints correspond to the MMFF structure and the reference DFT structure, respectively. The three examples are from the QM9 dataset. \revision{Atom colors follow the standard convention: C, gray; H, white; O, red; N, blue.}
}\label{fig:schematic_of_RDSM}
\end{figure}

\revision{To construct an energy-related manifold, we employed physics-informed internal coordinates ($q$-coordinates) to represent molecular structures, inspired by the geodesic interpolation method \cite{zhu2019geodesic} and Fukui's theory \cite{tachibana1979intrinsic}. By equipping the pull-back metric from this coordinate space, we obtained a Riemannian manifold $\mathcal{M}$ that encodes energetic relevance (details are provided in \cref{sec:construction of manifold,sec:manifold setting}).

The motivation for introducing such a manifold is closely tied to the formulation of DM. Conventional DMs perturb and denoise structures in Cartesian space, where the Euclidean metric does not directly correspond to the physical energy. Isotropic Gaussian noise in this space deforms rigid bonds and angles, displacing configurations far from low-energy basins; as a result, even an ideal denoising vector field can yield structures with substantial energy variance. By contrast, casting DM on a physics-informed manifold that aligns geometric distance with energy differences ensures that each perturbation and its subsequent denoising step are taken consistently on the PES, guiding structures toward the true minimum-energy basin.

Figure~\ref{fig:schematic_of_RDSM}a illustrates this contrast: in Euclidean space, geodesics deviate strongly from the energy minimization path, whereas in the physics-informed manifold they remain closely aligned with the PES. In the Euclidean space, energy level sets are distorted due to the anisotropic nature of interatomic interactions with respect to atomic Cartesian coordinates, such as bond stretching and angle bending. This distortion implies that isotropic noise on the Cartesian coordinates may result in highly unstable structures with significant energy deviations. As a result, there can be a large difference between the direction of the energy gradient and $\mathbf{x} - \tilde{\mathbf{x}}$. In contrast, the energy level sets in the Riemannian space form nearly concentric circles around $\mathbf{x}$, allowing the local geometry to encode the energy landscape. In this setting, the energy gradient and the geodesic vector field can be well aligned with each other. }

To demonstrate that the Riemannian manifold better captures the energetic properties of molecular structures, we analyzed the correlation between geodesic distances and energy differences. The geodesic distances were computed using the geodesic interpolation method, and the molecular energies were obtained by single-point DFT calculations. To assess the relationship between energy differences and structural mismatches, we used Pearson's correlation coefficient. The correlations were calculated for 1,000 randomly selected pairs of molecular structures, each optimized by the MMFF and DFT methods, respectively, from the QM9$_{\text{M}}$ dataset \cite{qm9m}. For comparison, we also considered the widely used root mean square deviation (RMSD) of atomic positions in Cartesian coordinates, as well as the mean absolute error of interatomic distances (D-MAE), shown in Supplementary Figure~4. The computational details of the measurements are given in \cref{sec:measurement_details}.

Figure~\ref{fig:schematic_of_RDSM}b presents a comparative analysis of the correlation between energy differences ($\abs{\Delta E}$) and the following two different measures of structural differences: RMSD and geodesic distance. The Pearson correlation coefficients, denoted as ``r'' in Figure~\ref{fig:schematic_of_RDSM}b, indicate the degree of the linear relationship between the energy difference and each structural deviation on a logarithmic scale. RMSD shows a lower correlation with the energy difference, with a value of 0.37, suggesting that it may not adequately capture energetically relevant structural deviations. In contrast, the geodesic distance exhibits a high correlation with a value of 0.90, indicating its effectiveness in capturing energy-associated structural changes. These results support the hypothesis that the physics-informed Riemannian manifold representation provides a more meaningful metric for assessing the molecular conformation on a PES.

Building on this insight, we introduce an R-DM framework, which operates on the designed Riemannian manifold. Hereafter, to clearly distinguish the conventional DM operating in Euclidean space from R-DM, we refer to the former as E-DM. For molecular structure optimization using E-DM and R-DM, we employ an algorithm based on an FM framework to refine sub-optimal structures to optimal ones (see \cref{section4-2}). Given an ideal denoising vector field, the ODE sampling transports the sub-optimal structure to the optimal one along the geodesic path. In E-DM, this path corresponds to the straight line, which is a linear interpolation between the structures in the Cartesian coordinate system. To illustrate the advantages of R-DM, Figure~\ref{fig:schematic_of_RDSM}c compares the energy profiles along the geodesic paths from the MMFF (sub-optimal) to DFT (optimal) structures, in both Euclidean and Riemannian spaces, together with the energy minimization paths (more examples are provided in Supplementary Figure~5). The energy profiles of the two geodesics differ substantially, even though the integrated path lengths of the trajectories are similar. While Euclidean geodesics show sharp energy increases, Riemannian ones decrease smoothly, mirroring energy minimization paths and highlighting the potential robustness of R-DM.

\subsection{Noise-sampling on Riemannian manifold}
To train DMs, noisy structures are generated by perturbing the original structures with controlled noise. In the E-DM method, a noisy structure $\tilde{\mathbf{x}}_\sigma$ is generated by adding noise to the original structure $\mathbf{x}$ as follows:
\begin{equation}
    \tilde{\mathbf{x}}_\sigma = \mathbf{x} + \sigma \boldsymbol{\xi},
\end{equation}
where $\sigma\in[0,\sigma_\mathrm{max}]$ is the noise scale and $\boldsymbol{\xi} \sim \mathcal{N}(\mathbf{0}, \mathbf{I}_{3N})$ is a Gaussian noise vector. In our approach, noisy structures are generated in a similar manner:
\begin{equation}
    \tilde{\mathbf{x}}_\sigma = \exp_{\mathbf{x}}{(\sigma \boldsymbol{\xi})},
\end{equation}
where $\mathbf{x}$ and $\tilde{\mathbf{x}}_\sigma \in \mathcal{M}$, and $\boldsymbol{\xi} \sim \mathcal{N}(\mathbf{0}, \mathbf{I}_d)$ is a tangent vector that represents the direction of the noise. In the process, the exponential map $\exp_{\mathbf{x}}$ transports a noise vector from the tangent space onto the manifold by moving along a geodesic starting from $\mathbf{x}$. This yields an exponential-wrapped normal distribution where the resulting noisy structures adhere to the underlying geometric constraints.

To compare the noisy structure distributions obtained from the two distinct spaces above, we sampled 1,000 noisy structures from each space and calculated their RMSD values and energy differences relative to $\mathbf{x}$. For reference, we also evaluated the MMFF structures corresponding to the same conformations as $\mathbf{x}$ as a baseline. The noise-level range was adjusted according to a predefined scheduling scheme, with the maximum $\sigma_\mathrm{max}$ set to ensure a distribution of RMSD similar to that of the MMFF structures (see Supplementary Figure~1 for scheduling scheme).


Figure~\ref{fig:noisy_distribution} shows a comparative analysis of molecular structures obtained by DFT, MMFF, Euclidean noise-sampling, and Riemannian noise-sampling, presenting (a) example structures of three molecules with their RMSD and absolute energy difference ($\abs{\Delta E}$) relative to the DFT references, and (b) the distributions of RMSD and $\abs{\Delta E}$ for 1,000 randomly sampled structures from each method.

The results indicate that while the RMSD distributions of the Riemannian and Euclidean noise-sampling are similar, the $\abs{\Delta E}$ distributions show a substantial discrepancy. Specifically, the Euclidean noise causes greater energetic instability, often involving new bond formation or breaking. 
In contrast, both the $\abs{\Delta E}$ distributions of the Riemannian and MMFF cases show much smaller energy changes with intact molecular structures (no bond formation or breaking). In both cases, primary structural differences occur in the bond and dihedral angles with respect to the reference structures, as shown in Figure~\ref{fig:noisy_distribution}a. Furthermore, this suggests that the structural perturbation on the Riemannian manifold defined by \cref{eq:metric} can approximate those on the respective PES (see Supplementary Figure~6 for more detailed results). 

E-DM and R-DM are trained to denoise structures generated by the Euclidean and Riemannian noise-sampling, respectively. In contrast to E-DM, where Gaussian noise in Euclidean space directly serves as a tractable initial distribution, the analogous one in R-DM is intractable \cite{chavel1984eigenvalues}. However, our findings indicate that the Riemannian noise-sampled structures consistently remain within the same potential well as the original molecule, thereby enabling the use of easily obtainable structures, such as those obtained via MMFF, as the initial structure. \Cref{sec:Revised Riemannian noise model,sec:revised training objective,section4-2} provides more details of the noising process including the formulation of R-DM in perspective of the SM and FM frameworks, the corresponding sampling algorithms and training framework.

\subsection{Performance on molecular structure optimization}\label{subsec2-3}
We first evaluate the performance of R-DM on the QM9 dataset, using conventional E-DM as a baseline.
To assess the precision in molecular structure optimization, the MMFF-optimized structures of the reference molecules for each conformation were used as the initial structures. 
For optimized structures, we compare measurements that can be calculated directly from molecular structures, such as RMSD and D-MAE, together with the energy difference $\abs{\Delta E}$. These metrics are calculated relative to the reference DFT results, and $\abs{\Delta E}$ is obtained by single-point DFT calculations.


\Cref{tab:tab1} shows mean and median values for each measurement for the molecules in the test set, while the distributions of RMSD and $\abs{\Delta E}$ are illustrated in \Cref{figure3}. Additional distributions, including those for D-MAE and $L^2$ norm in $q$-coordinates, can be found in Supplementary Figure~7. Both E-DM and R-DM refined the initial MMFF structures substantially, while R-DM achieved much higher accuracy. It is important to note that the following improvements reflect refinement capabilities of the models rather than direct performance comparisons with MMFF, which is not tailored to DFT data and serves as an initialization method. In particular, the median improvement in $\abs{\Delta E}$ was greater than 20$\times$ for R-DM and about 5$\times$ for E-DM, while the median improvements in RMSD and D-MAE were less different between the two methods. The substantial decrease of $\abs{\Delta E}$ in R-DM can be attributed to the denoising process in the physics-informed Riemannian metric, which is highly correlated with the energy change, as described above. Moreover, R-DM achieved chemical accuracy, an error below 1 kcal/mol, in the mean value of $\abs{\Delta E}$, highlighting the model's robustness in molecular structure optimization.


Building on R-DM's accurate predictions, our method facilitates further refinement through post-optimization with DFT at substantially reduced computational cost compared to conventional approaches. \Cref{tab:post_opt} compares the computational costs of this post-optimization in terms of the number of force calls, self-consistent field (SCF) cycles and times. Specifically, a force call corresponds to one iteration of the optimization process, and each force call entails iterative SCF computations.

R-DM achieved a remarkable improvement by reducing the number of iterations and the computation time by more than twofold compared to MMFF. In most cases, R-DM obtained the optimized structure in just three force calls, highlighting its efficiency in the post-optimization process. Moreover, the computational overhead of R-DM itself is minimal. Although denoising approaches inherently require multiple steps, R-DM achieves sufficient accuracy in only 0.55 seconds for 128 steps or 0.07 seconds for 16 steps\footnote{Measured with 16 threads on Intel Xeon Gold 6326 CPUs at 2.90 GHz}, a cost negligible compared to subsequent DFT calculations (see Supplementary Table~4).


To further validate R-DM, we evaluated it on the QM7-X dataset \cite{qm7x}, which provides non-equilibrium molecular structures generated by perturbations along vibrational modes. Such perturbations offer a complementary scenario that resembles the noisy configurations encountered during the diffusion process on the PES. Firstly, we conducted an analysis consistent with that performed on QM9$_\text{M}$, comparing the real potential energy surface and the Riemannian manifold imitating their physics. We examined the correlation between the metric distances and energy differences\footnote{Following previous studies \cite{qm7x, morered}, molecular energies were calculated using the PBE0+MBD level of theory.}, and analyzed the distributional similarities (see Supplementary Figures~8 and 9). These comparisons revealed patterns closely aligned with those observed in QM9$_\text{M}$, thereby reinforcing the generality of our approach.

\Cref{tab:qm7x_results} summarizes the optimization results on the QM7-X test set, where non-equilibrium structures were used as initial configurations for E-DM, R-DM, and MoreRed-JT \cite{morered} methods. Due to cases that occasionally led to convergence failures, we report median values rather than mean values to provide more robust statistics. Relative to the baseline E-DM, R-DM consistently improves all measurements, achieving a median absolute energy error of 0.277 kcal/mol. Compared to MoreRed-JT, R-DM shows slight gains in structural measurements (RMSD and D-MAE), while delivering comparable energy accuracy.


Despite R-DM's substantial reduction in structural deviations, its improvement in energy accuracy is relatively modest compared to the results on QM9. Further analysis revealed that some energy errors can persist even when the $L^2$ norm in $q$-coordinates is small. This is particularly pronounced for reference structures with dihedral angles near 0\textdegree{} or 180\textdegree{}, where even slight angular deviations can lead to large changes in energy. Consequently, the absolute energy errors in R-DM exhibit a broader distribution. For example, while the second-quartile (median) values for R-DM and MoreRed-JT are similar, the first-quartile energy error of R-DM is nearly half that of MoreRed-JT. These trends are clearly illustrated in Supplementary Figure~10, which presents box plots for various evaluation metrics, including absolute energy errors. This observation suggests that Zhu's metric, while effective in modeling local structural variations, may fall short in capturing energy sensitivity related to specific conformational features such as dihedral torsions. Accordingly, there is a room for improving R-DM's performance through further refinement of the underlying Riemannian metric.

The more pronounced manifestation of this issue in QM7-X compared to QM9$_\text{M}$ can be attributed to differences in the initial structures; MMFF-optimized structures in QM9$_\text{M}$ typically exhibit smaller dihedral angle deviations, whereas QM7-X deliberately introduces perturbations in these torsionally sensitive coordinates. When using the MoreRed-JT results as initial structures, R-DM achieves notable performance across all measurements, attaining energy accuracy comparable to that observed in the QM9 experiments. The hyperparameter settings and sampling options for E-DM and R-DM are detailed in the Supplementary Section~1.

\revision{We further evaluated R-DM on conformer generation and ensemble property prediction tasks using the GEOM-QM9 dataset. By refining diverse initial conformations obtained by methods such as ETKDG and generative models, R-DM achieved improved coverage and matching scores as well as more accurate ensemble property predictions. These findings indicate that, although R-DM is designed as an optimization model, its ability to reliably recover accurate low-energy structures makes it broadly useful for tasks where structural fidelity is essential.
The detailed results and discussion are provided in Supplementary Section~6.}

\section{Discussion}
The development of a Riemannian denoising model (R-DM) was motivated by the need for a model with the so-called chemical accuracy for molecular modeling. We first considered what constitutes an accurate molecular structure; accurate in what sense? Our answer was that the accuracy should be assessed in terms of energy. We recognized that conventional metrics like RMSD do not effectively capture energetic changes essential to achieving the chemical accuracy. To address this, we proposed a physics-informed Riemannian space for the noising/denoising process, ensuring a strong correlation between spatial distance and energy variation.

\revision{To use R-DM for molecular structure optimization, we trained a vector field under a unified view of score matching and flow matching paradigms. This formulation enables the model to be applied during inference without dependence on predefined noise schedules or the choice of initial structures, making it broadly applicable to optimization tasks. In this light, it is instructive to compare R-DM with machine learning force fields (MLFFs), another family of models for molecular structure optimization. MLFFs aim to approximate the potential energy surface itself by learning forces derived from \emph{ab initio} computation, which makes them powerful for molecular dynamics and iterative optimization. In contrast, R-DM does not attempt to learn the PES explicitly but instead incorporates a physics-informed manifold as an inductive bias for learning the denoising vector field. This distinction highlights the complementary nature of the two approaches: MLFFs directly emulate physical forces, whereas R-DM leverages geometric structure to achieve data-efficient optimization.}

\revision{A key challenge of R-DM is its intractable prior on the manifold, which we have addressed empirically by using easily obtainable initial structures. However, our current results have not been extensively validated across a wide range of molecular systems. In particular, our current validation focused primarily on small molecules due to the computational cost of quantum chemical calculations, though our approach itself is not inherently restricted by molecular size. Future research will aim to enhance the method's robustness and utility for a wide range of molecular modeling tasks. Furthermore, while our evaluations have focused on equilibrium structures, the underlying principle of energy-centric modeling can be extended to a wider range of molecular simulations where precise energy calculations are crucial. Further advancements can also be achieved by refining the Riemannian metric or adapting it to specific tasks.}

\section{Methods}

\subsection{Construction of the physics-informed Riemannian manifold}\label{sec:construction of manifold}
In ordinary Euclidean space, the metric is the same everywhere, so distances are measured uniformly and the shortest path is always a straight line. In contrast, a Riemannian manifold carries a smoothly varying metric tensor that depends on the position, enabling the local geometry to naturally encode anisotropic features of the space and allowing geodesics to curve.

In molecular structure modeling, a configuration is more naturally described in internal coordinates, such as bond lengths, bond angles, and dihedral angles, than in raw Cartesian positions. It is extended to a general coordinate and space, requiring a position-dependent metric and thus the concept of a Riemannian manifold. This geometric perspective is more elaborated with Fukui's intrinsic reaction coordinate (IRC) theory, which describes molecular transitions occurring through steepest-descent trajectories on the mass-weighted potential energy surface. The trajectories naturally endow the general coordinate space with a distinctive metric tensor, commonly called Fukui's metric; consequently, the transition paths emerge as geodesics on the resulting Fukui's manifold \cite{tachibana1979intrinsic}.

For the design of the energy-related manifold, we employed physics-informed internal coordinates to represent molecular structures, inspired by the geodesic interpolation method \cite{zhu2019geodesic}, in which a molecule with $N$ atoms is represented by a vector of interatomic terms, $\mathbf{q}\in\mathbb{R}^{d}$. The Cartesian coordinate $\mathbf{x}\in\mathbb{R}^{3N}$ is related to the internal coordinates by the smooth and locally invertible map
\begin{equation}\label{eq:internal-coordinate embedding}
    \Phi:\;\mathbf{x}\;\mapsto\;\mathbf{q}=\Phi(\mathbf{x}).
\end{equation}
Thus, any molecular structure can be expressed either in Cartesian form $\mathbf{x}$ or, equivalently, in $q$-coordinates $(q^{e})_{e=1,\dots,d}$. The $e$-th component of $\mathbf{q}=(q^{1},\dots,q^{d})$ is defined as
\begin{align}
    \label{eq:q-coordinate definition}
    q^{e} &= \exp\!\left[-\alpha\,(r^{e}-1)\right] + \beta\,\frac{1}{r^{e}},\\
    r^{e} &= \frac{\|\mathbf{x}_{i}-\mathbf{x}_{j}\|_{2}}{R_{i}+R_{j}},
\end{align}
where $i$ and $j$ denote the atomic indices associated with the interatomic index $e$, $\mathbf{x}_{i}$ is the Cartesian position of the $i$-th atom, and $R_{i}$ is its covalent radius. The parameters are set to $\alpha=1.7$ and $\beta=0.01$ following Zhu \textit{et al.}\ \cite{zhu2019geodesic}.  

The set of points $\mathbf{q}$ corresponding to feasible molecular structures forms a smooth sub-manifold immersed in $\mathbb{R}^{d}$ \cite{nash1956imbedding}. While the choice of coordinates labels each structure, quantitative comparison requires a Riemannian metric. We endow the configuration space with the pull-back metric following the Zhu \textit{et al.}\ \cite{zhu2019geodesic}
\begin{equation}\label{eq:metric}
    g_{ij} \;=\; \sum_{e=1}^d\frac{\partial q^{e}}{\partial x^{i}}\,
                 \frac{\partial q^{e}}{\partial x^{j}}, 
    \qquad ds^{2}=g_{ij}\,dx^{i}\,dx^{j},
\end{equation}
where $x^i$ is a Cartesian coordinate, and the metric obtained by pulling back the Euclidean metric of $q$-space through $\Phi$. If this metric further approximates the Fukui metric derived from the PES,
\begin{equation}\label{eq:Fukui metric}
    g_{ij}\;\approx\; \frac{\partial E}{\partial x^{i}}\,
                      \frac{\partial E}{\partial x^{j}},
\end{equation}
then the geodesic distance between two structures within the same conformational well becomes directly proportional to their energy $E$ difference \cite{tachibana1979intrinsic}. Consequently, shortest paths on the manifold correspond to energetically minimal deformations, providing a physically meaningful notion of structural similarity. These aspects are well embedded in the internal coordinate design; the exponential decay of $q^e$ with interatomic distance ratio $r^e$ emphasizes short-range steric repulsion between atoms while rendering long-range displacements weakly influential, thus faithfully encoding interatomic physics within the manifold.

\subsection{Manifold setting}\label{sec:manifold setting}
This subsection specifies computational details about the Riemannian geometry used in the R-DM framework, covering the construction of the metric tensor, the representation of tangent spaces, and the numerical solution of geodesics required by the exponential map.

\subsubsection{Metric Tensor}
The Riemannian metric in \cref{eq:metric} is obtained by pulling back the standard Euclidean metric on \(\mathbb{R}^{d}\):
\begin{equation}\label{eq:metric tensor definition}
  g(\mathbf{x})
  \;=\;
  \mathbf{J}_{\Phi}(\mathbf{x})^{\!\top}\,
  \mathbf{J}_{\Phi}(\mathbf{x}),
  \qquad
  \mathbf{J}_{\Phi}(\mathbf{x})
  =\frac{\partial\Phi}{\partial\mathbf{x}}\bigg\rvert_{\mathbf{x}}
  \in\mathbb{R}^{d\times 3N},
\end{equation}
where all Jacobian entries admit analytical expressions $\partial q^{e}/\partial x^i$. While $\mathbf{q}$ merely labels a point on the manifold, the metric $g(\mathbf{x})$ endows that point with local geometric structures, such as distances, angles, and volume elements, independent of the chosen coordinates.

\subsubsection{Tangent Space and Gaussian Sampling}\label{sup:tangent}
For any Cartesian configuration $\mathbf{x}\in\mathbb{R}^{3N}$ the pull-back metric endows the ambient space with an inner product $\langle\!\boldsymbol{\delta x},\boldsymbol{\delta x}'\!\rangle_{g(\mathbf{x})} =\boldsymbol{\delta x}^{\!\top} g(\mathbf{x})\,\boldsymbol{\delta x}'$. Hence, $T_{\mathbf{x}}\mathcal{M}\cong\mathbb{R}^{3N}$. Since rigid translations and rotations provide six redundant directions, moving to the internal coordinate image $\mathcal{M}_{\mathrm{int}}=\operatorname{Im}\Phi$ reduces the effective dimension to $3N-6$. All computations are therefore performed in $\mathcal{M}_{\mathrm{int}}$, which excludes these roto-translational degrees of freedom.

\paragraph{Column-space characterization.}
The column space of $\mathbf{J}_\Phi(\mathbf{x})$
\begin{equation}
    \operatorname{Col}\!\bigl(\mathbf{J}_\Phi(\mathbf{x})\bigr)
    \;=\;
    \operatorname{Im}\!\bigl[D\Phi_{\mathbf{x}}\bigr]
    \;=\;
    T_{\mathbf{q}}\mathcal{M}_{\mathrm{int}},
    \qquad
    \mathbf{q}=\Phi(\mathbf{x}),
\end{equation}
is precisely the tangent space of the embedded manifold $\mathcal{M}_\mathrm{int}\subset\mathbb{R}^{d}$. Based on the definitions of $\mathcal{M}$ and $\mathcal{M}_{\mathrm{int}}$, the differential $D\Phi_{\mathbf{x}}$ establishes a one-to-one correspondence between the $(3N-6)$-dimensional subspace orthogonal to rigid-body motions and $T_{\mathbf{q}}\mathcal{M}_{\mathrm{int}}$:
\begin{equation}
  D\Phi_{\mathbf{x}}: T_{\mathbf{x}}\!\mathcal{M} \!\longrightarrow\! T_{\mathbf{q}}\mathcal{M}_{\mathrm{int}}. 
\end{equation}

\paragraph{Gaussian sampling in the tangent space.}
To simulate Brownian motion on the manifold and generate noisy structure, it is required to sample tangent vectors from the Gaussian distribution on the tangent space $\boldsymbol{\xi}_{\mathbf{x}}\in T_{\mathbf{x}}\!\mathcal{M}$ that are isotropic with respect to the metric $g(\mathbf{x})$. In practice, we sample the vector in $T_{\mathbf{q}}\!\mathcal{M}_\mathrm{int}$ first, and then map it back to the $T_\mathbf{x}\!\mathcal{M}$. More specifically, we first draw $\mathbf{z}\sim\mathcal{N}(\mathbf{0},\mathbf{I}_{d})$ in $\mathbb{R}^{d}$ and project it onto the tangent space $T_{\mathbf{q}}\!\mathcal{M}_\mathrm{int}$ by:
\begin{equation}
    \boldsymbol{\xi}_{\mathbf{q}}=\mathbf{P}_{\!J}\,\mathbf{z}, 
    \qquad \mathbf{P}_{\!J} =\mathbf{J}_\Phi (\,\mathbf{J}_\Phi^{\!\top}\mathbf{J}_\Phi)^{-1} \mathbf{J}_\Phi^{\!\top}.
\end{equation}
Successively, map the projected vector to $T_{\mathbf{x}}\!\mathcal{M}$ via the pseudoinverse $\mathbf{J}_\Phi^{+} =\mathbf{J}_\Phi^{\!\top} (\mathbf{J}_\Phi\mathbf{J}_\Phi^{\!\top})^{-1}$:
\begin{equation}
    \boldsymbol{\xi}_{\mathbf{x}} =\mathbf{J}_\Phi^{+}\,\boldsymbol{\xi}_{\mathbf{q}} 
    =\mathbf{J}_\Phi^{\!\top} (\mathbf{J}_\Phi\mathbf{J}_\Phi^{\!\top})^{-1} \mathbf{P}_{\!J}\,\mathbf{z}.
\end{equation}
One can verify ${\mathbb{E}}\!\bigl[\boldsymbol{\xi}_{\mathbf{x}} \boldsymbol{\xi}_{\mathbf{x}}^{\!\top}\bigr] =g(\mathbf{x})^{-1}$, ensuring metric-isotropic noise. For scaling, we simply multiply a scale factor $\alpha$ to the $\boldsymbol{\xi}_{\mathbf{x}}$.

\subsubsection{Geodesic Equation and Exponential Map}
Unlike many manifold simulators that integrate geodesics in the internal coordinate chart, we solve the second–order ODE directly in Cartesian space $\mathbf{x}\in\mathbb{R}^{3N}$:
\begin{equation}\label{eq:geo-x}
  \ddot{x}^{k}+ \Gamma^{k}_{ij}(\mathbf{x}) \dot{x}^{i}\dot{x}^{j} = 0, \qquad i,j,k = 1,\dots,3N,
\end{equation}
where the Christoffel symbols $\Gamma^{k}_{ij}$ are expressed in terms of the distance–coordinate map $\Phi$.

The exponential map converts a tangent vector into a point on the manifold by shooting a geodesic for unit time. Formally, for $\mathbf{x}\in\mathcal{M}$ and $\boldsymbol{\xi}\in T_{\mathbf{x}}\!\mathcal{M}$ let $\gamma_{\mathbf{x},\boldsymbol{\xi}}:[0,1]\!\to\!\mathcal{M}$ satisfy the geodesic ODE in \cref{eq:geo-x} with $\gamma(0)=\mathbf{x}$ and $\dot{\gamma}(0)=\boldsymbol{\xi}$. The exponential map at~$\mathbf{x}$ is then $\exp_{\mathbf{x}}(\boldsymbol{\xi})= \gamma_{\mathbf{x},\boldsymbol{\xi}}(1)$. Geometrically, $\exp_{\mathbf{x}}$ transports the molecule from $\mathbf{x}$ along the direction of $\boldsymbol{\xi}$ while preserving the shortest-path property of a geodesic. In a sufficiently small neighborhood, the map is a diffeomorphism and can be approximated by a linear map (locally Euclidean).

\paragraph{Christoffel computation via Hessian--Jacobian contraction.}
Let $q^{e}\!=\!\Phi^{e}(\mathbf{x})$ denote the $e$-th internal coordinate $(e=1,\dots,d)$. The chain rule yields
\begin{equation}\nonumber
  g_{ij}=\sum_{e}\partial_{i}q^{e}\,\partial_{j}q^{e},
\end{equation}
so that
\begin{equation}\nonumber
  \partial_{k}g_{ij}
  =2\sum_{e}\partial_{k}\partial_{i}q^{e}\,\partial_{j}q^{e}.
\end{equation}
Plugging this into the usual definition of the Christoffel symbol (with the Levi-Civita connection)
\begin{equation}\nonumber
    \Gamma^{k}_{ij} =\frac{1}{2} g^{k\ell} (\partial_{i}g_{j\ell} +\partial_{j}g_{i\ell} -\partial_{\ell}g_{ij})
\end{equation}
one avoids an explicit inverse of the $(3N)\!\times\!(3N)$ metric by contracting the Hessian of $\Phi$ with the pseudoinverse of its Jacobian:
\begin{equation}\label{eq:christoffel in x}
\Gamma^{k}_{ij} =
  \bigl(\mathbf{J}_{\Phi}^{+}\bigr)_{k e}\,
  \frac{\partial^{2}q^{e}}{\partial x^{i}\partial x^{j}}.
\end{equation}
In practice, the Hessian and Jacobian are computed analytically utilizing sparse operations, while the pseudoinverse is computed numerically.

\paragraph{Time integration.}
Equation \eqref{eq:geo-x} is initially integrated with the Heun method. Because the geodesic equation preserves the magnitude of the velocity, we monitor the relative error in that norm; if the accumulated error exceeds a prescribed threshold, we recompute the trajectory using a fourth-order Runge–Kutta (RK4) scheme. At each integration step, we evaluate the Christoffel symbols in Eq.~\eqref{eq:christoffel in x} and update $\mathbf{x}$ and $\dot{\mathbf{x}}$ as
\begin{equation}
  \mathbf{x}\leftarrow \mathbf{x}+ \dot{\mathbf{x}}\,\Delta t,\;
  \dot{\mathbf{x}}\leftarrow
  \dot{\mathbf{x}}-\Gamma(\mathbf{x})[\dot{\mathbf{x}},\dot{\mathbf{x}}]\,
  \Delta t,
\end{equation}
where $\Gamma[\dot{\mathbf{x}},\dot{\mathbf{x}}]$ denotes the contraction in Eq.~\ref{eq:geo-x}.

\subsection{Riemannian noise model}\label{sec:Revised Riemannian noise model}
Both the SM and FM frameworks rely on synthetic noisy structures; SM regards them as time marginals of a diffusion process, whereas FM interprets them as points along a transport curve. In this subsection, we demonstrate that our noising scheme satisfies both views simultaneously—acting as the manifold heat kernel in the diffusion perspective and as a geodesic interpolant in the transport perspective.

We generate noisy structures by drawing an isotropic Gaussian vector $\boldsymbol{\xi}\sim \mathcal{N}(\mathbf{0}, \mathbf{I})$ in the tangent space of each clean conformation $\mathbf{x}$ and mapping it back to the manifold with the exponential map,
\begin{equation}\label{eq:revised noise sample generator}
    \tilde{\mathbf{x}}_\sigma = \exp_{\mathbf{x}}(\sigma \boldsymbol{\xi}), \qquad 0\le \sigma\le \sigma_\mathrm{max}.
\end{equation}
When the manifold is flat (Euclidean), the exponential map reduces to the familiar additive form $\tilde{\mathbf{x}}_\sigma=\mathbf{x}+\sigma\boldsymbol{\xi}$. The noise scale $\sigma$ is bounded by $\sigma_\mathrm{max}$, which is set to be small enough so that each perturbation $\tilde{\mathbf{x}}_\sigma$ stays within the same conformational basin. Because the exponential map follows a geodesic prescribed by the physics-informed metric, bond lengths and angles stay chemically reasonable even for the largest $\sigma_\mathrm{max}$.

The distribution of $\tilde{\mathbf{x}}_\sigma$ is exponential-wrapped Gaussian centered at $\mathbf{x}$; for small $\sigma$ it closely approximates the manifold heat-kernel \cite{varadhan1966asymptotic}, which underpins the diffusion perspective. The corresponding diffusion process $\{\mathbf{X}_t\}_{t\in[0,1]}$ on the manifold follows the stochastic differential equation
\begin{equation}\label{eq:revised forward sde}
    d\mathbf{X}_t=\sqrt{\frac{d\sigma^2(t)}{dt}}d\mathbf{B_t^{\mathcal{M}}}, \qquad \sigma(t)=\kappa(t)\sigma_\mathrm{max},\qquad \mathbf{X}_0\sim p_\mathrm{data}(\mathbf{x})
\end{equation}
where $\mathbf{B}^\mathcal{M}_t$ is standard Brownian motion on the manifold and the increasing function $\kappa:[0,1]\to[0,1]$ is time-parameterization. Note that $1/\kappa^2(t)$ is proportional to the signal-to-noise ratio. With this parameterization, each time marginal $\mathbf{X}_t\sim p_t$ corresponds to $\tilde{\mathbf{x}}_{\sigma(t)}$ defined in \cref{eq:revised noise sample generator}.

From the FM perspective, we view denoising as transporting a maximally corrupted structure $\tilde{\mathbf{x}}$ back to its clean counterpart $\mathbf{x}$. Here, $\exp_{\mathbf{x}}^{-1}(\tilde{\mathbf{x}})=\sigma_\mathrm{max}\boldsymbol{\xi}$ is the tangent vector connecting them, and the transport curve is given by $\mathbf{x}_t = \exp_{\mathbf{x}}\big(\sigma(t)\boldsymbol{\xi}\big)$, where $\sigma(t)=\kappa(t)\sigma_\mathrm{max}$ and $\kappa$ serves as the interpolation ratio between $\mathbf{x}$ and $\tilde{\mathbf{x}}$. Note that the direction of time here is the reverse of the conventional FM setting, as $t=0$ corresponds to the clean structure and $t=1$ to the maximally corrupted one. By randomly sampling pairs $(\mathbf{x},\tilde{\mathbf{x}})$ following \cref{eq:revised noise sample generator}, the time marginals of the geodesic interpolant $\mathbf{x}_t$ coincide with those of the diffusion process $p_t$. Consequently, the same noise construction supplies both the heat-kernel needed for the SM framework and the geodesic transports needed for the FM framework; the two training paradigms are unified by a common Riemannian noise model.

\subsection{Unified denoising objective}\label{sec:revised training objective}
We aim to construct a denoising network, which is a vector field on the manifold that, at any noisy structure $\tilde{\mathbf{x}}_\sigma$, points toward the corresponding clean structure $\mathbf{x}$. In this section, we present the training objectives for both the SM and FM frameworks, which share the same functional form and differ only in the choice of loss-weight. We then introduce our reparameterization scheme and the following unified training objective.

From the SM viewpoint, the denoising model acts as a score function estimator $\mathbf{s}_\theta(\mathbf{x}_t, t)\approx\nabla \log p_{t}(\mathbf{x}_t)$. Using Varadhan's asymptotics, we obtain $\nabla_{\mathbf{x}_t}\log p_{t\mid0}(\mathbf{x}_t \mid \mathbf{x}_0)\approx \exp_{\mathbf{x}_t}^{-1}(\mathbf{x}_0)/\sigma^2(t)$. Accordingly, the denoising model can be trained using the standard score matching objective
\begin{equation}\label{eq:loss-DSM}
    \mathcal{L}_{\mathrm{SM}}= \mathbb{E}_{t, \mathbf{x}_0,\mathbf{x}_t}\biggl[
    w(t)\lVert \mathbf{s}_\theta(\mathbf{x}_t, t) -  \exp_{\mathbf{x}_t}^{-1}(\mathbf{x}_0)/\sigma^2(t)
    \rVert^2_g\biggr],
\end{equation}
where $w(t)$ is a loss-weight term for training efficiency.

In the FM framework, the denoising model estimates the velocity field $\mathbf{v}_\theta(\mathbf{x}_t, t)\approx\mathbf{u}_t(\mathbf{x})$, where the field is the time derivative of diffeomorphism (\textit{flow}) $\psi:[0,1]\times \mathcal{M}\to \mathcal{M}$ transporting the probability density $\mathbf{x}_t \sim p_t$ as follows, 
\begin{equation}\label{eq:flow formula}
    p_t=[\psi_t]_{\sharp}p_1, \qquad \frac{d}{dt}\psi_t(\mathbf{x})=\mathbf{u}_t(\psi_t(\mathbf{x})).
\end{equation}
Here, $\sharp$ denotes the push-forward operator. Following the conditional FM formula \cite{chen2023riemannian}, we have the $\mathbf{x}_0$ conditional transport velocity $\mathbf{u}_{t}(\mathbf{x}_t\mid\mathbf{x}_0)=-\tfrac{\dot{\sigma}}{\sigma}(t)\exp_{\mathbf{x}_t}^{-1}(\mathbf{x}_0)$, and the FM loss for the denoising model is written as
\begin{equation}\label{eq:loss-FM}
    \mathcal{L}_{\mathrm{FM}}=\mathbb{E}_{t, \mathbf{x}_0, \mathbf{x}_t} \left[
    w(t)\lVert \mathbf{v}_\theta(\mathbf{x}_t,t) + \exp_{\mathbf{x}_t}^{-1}(\mathbf{x}_0)\frac{\dot{\sigma}}{\sigma}(t)
    \rVert_g^2
    \right].
\end{equation}

Both $\mathbf{u}_t(\cdot\mid\mathbf{x}_0)$ and $\nabla \log p_{t\mid0}(\cdot \mid \mathbf{x}_0)$ are proportional to the tangent vector that points from the current noisy structure toward the clean one. In this light, we reparameterize both denoising models in terms of a single time-independent vector field $\mathbf{f}_\theta:\mathcal{M}\to T\mathcal{M}$, setting
\begin{equation}\label{eq:reparameterization}
 \mathbf{s}_\theta(\mathbf{x},t):=\frac{1}{\sigma^2}(t)\mathbf{f}_\theta(\mathbf{x}), \quad \mathbf{v}_\theta(\mathbf{x},t):=-\frac{\dot{\sigma}}{\sigma}(t)\mathbf{f}_\theta(\mathbf{x}).
\end{equation}
We can rewrite training objectives in \cref{eq:loss-DSM,eq:loss-FM} in a single unified form that differs only in the choice of loss-weight $w$:
\begin{equation}\label{eq:unified loss}
    \mathcal{L}_{\mathrm{uni}}=
    \mathbb{E}_{t, \mathbf{x},\tilde{\mathbf{x}}_{\sigma(t)}}
    \biggl[w(t)\lVert 
    \mathbf{f}_\theta(\tilde{\mathbf{x}}_{\sigma(t)}) - \exp_{\tilde{\mathbf{x}}_{\sigma(t)}}^{-1}(\mathbf{x})
    \rVert_g^2\biggr].
\end{equation}
In practice, we sampled the noisy structure $\tilde{\mathbf{x}}_{\sigma(t)}$ from the clean counterpart $\mathbf{x}$ following \cref{eq:revised noise sample generator} with uniformly sampled time $t\in[0, 1]$. The loss weight $w(t)$ is determined as $\tfrac{1}{\sigma^2}(t)$ to normalize the size of the loss, and the profile of $\sigma(t)$ is presented in Supplementary Figure~1.

\subsection{Sampling algorithms}\label{section4-2}
Once the vector field $\mathbf{f}_\theta$ is trained, we can perform inference in two ways: a \textit{diffusion-style} sampler based on the time-reversed SDE, or a \textit{FM-style} sampler based on an ODE along the geodesic curve.

For the diffusion case, we need to simulate the time-reversed SDE
\begin{equation}\label{eq:time-reversed SDE}
    d\mathbf{X}_t = -\frac{d\sigma^2(t)}{dt}\mathbf{s}_\theta (\mathbf{X}_t,t) dt + \sqrt{\frac{d\sigma^2(t)}{dt}} d\tilde{\mathbf{B}}_{t}^{\mathcal{M}}, \qquad \mathbf{s}_\theta(\mathbf{x}, t)=\frac{\mathbf{f}_\theta(\mathbf{x})}{\sigma^2(t)},
\end{equation}
where $\tilde{\mathbf{B}}_t^{\mathcal M}$ is Brownian motion on the manifold. Following Bortoli \textit{et al.} \cite{de2022riemannian}, we discretize the SDE with a geodesic random walk; at every step a tangent-space increment is drawn and mapped back with the exponential map.

For FM sampling, we solve the ODE
\begin{equation}\label{eq:FM ODE}
    \frac{d\mathbf{x}_t}{dt} = \mathbf{v}_\theta(\mathbf{x}_t, t), 
    \qquad \mathbf{v}_\theta(\mathbf{x},t) = -\frac{\dot{\sigma}(t)}{\sigma(t)}\,\mathbf{f}_\theta(\mathbf{x}).
\end{equation}
In our parameterization, the target distribution corresponds to $t=0$ and the source distribution to $t=1$, so the ODE is integrated backward in time. We discretize the ODE using a first-order Euler scheme, mapping each tangent update back to the manifold via the exponential map.

Consequently, a single trained network $\mathbf{f}_\theta$ supports both diffusion-style inference via $\mathbf{s}_\theta$ as in the \cref{eq:time-reversed SDE} and FM inference via $\mathbf{v}_\theta$ as in the \cref{eq:FM ODE}, with no additional training or parameter tuning. We adopt FM-style for all main results because it yields slightly lower errors than its diffusion-style counterparts while keeping the number of function evaluations identical. Comparative results for the sampling algorithms, along with the effect of the number of function evaluations on accuracy and convergence, are provided in the Supplementary Section~2.

\subsection{Implementation and training details}
To train the R-DM model efficiently, we adopted a two-step approach. Initially, R-DM was pretrained using noisy structures generated through Euclidean noise-sampling, which can be obtained immediately. This was followed by fine-tuning using noisy structures obtained from Riemannian noise-sampling. Both stages employ the same underlying noise scheduler $\hat{\sigma}(t)$ but with different effective ranges. Instead of truncating the time domain, we rescale the scheduler so that $t \in [0,1]$ is preserved; for example, $\sigma(t) = \hat{\sigma}\!\left(t_\mathrm{max} t\right)$. The cutoff parameter $t_\mathrm{max}$ was chosen by matching the noisy distributions to the MMFF ensemble in terms of RMSD and D-MAE (see \revision{Supplementary Figure~6}), resulting in $t_\mathrm{max}=0.3$ for pretraining and $t_\mathrm{max}=0.03$ for fine-tuning. For E-DM we use the full range with $t_\mathrm{max}=1$. The underlying noise-scheduler $\hat{\sigma}(t)$ and $\hat{\beta}(t)$ profile, which induced $\sigma(t)$ and $\beta(t)$, is plotted in \revision{Supplementary Figure~1}.

During pretraining, we approximate the logarithmic map $\exp^{-1}_{\tilde{\mathbf{x}}_\sigma}(\mathbf{x})$ for computational efficiency as follows:
\begin{equation}
    \exp_{\tilde{\mathbf{x}}_\sigma}^{-1}(\mathbf{x})\approx \mathbf{J}_\Phi^{+} \underbrace{\frac{\mathbf{P}_J \Delta \mathbf{q}}{\lVert\mathbf{P}_J \Delta \mathbf{q}\rVert_2} \lVert\Delta\mathbf{q}\rVert_2}_{\approx \exp_{\tilde{\mathbf{q}}_\sigma}^{-1}(\mathbf{q})}
\end{equation}
where $\Delta \mathbf{q}:=\Phi(\mathbf{x}) -\Phi(\tilde{\mathbf{x}}_\sigma)$ denotes the difference between two structures in $q$-coordinate in $\mathbb{R}^d$, and $\mathbf{P}_J$ represents the projection of $\Delta \mathbf{q}$ onto $T_{\tilde{\mathbf{q}}_\sigma}\!\mathcal{M}_{\mathrm{int}}$. The size relationship among the three vectors is as follows: $\| \exp_{\tilde{\mathbf{q}}_\sigma}^{-1}(\mathbf{q}) \|_g \geq \| \Delta \mathbf{q} \|_2 \geq \| \mathbf{P}_J\Delta \mathbf{q} \|_g$. To correct the reduction in length due to projection, we rescale the vector to match $\| \Delta \mathbf{q} \|_2$. Accurately computing $\exp_{\tilde{\mathbf{x}}_\sigma}^{-1}(\mathbf{x})$ involves a non-negligible computational cost, which entails a computational overhead determining the geodesic path between $\mathbf{x}$ and $\tilde{\mathbf{x}}_\sigma$ on the manifold and obtaining the associated tangent vector. For efficiency, this approximation is applied instead. During fine-tuning, the noisy structures $\tilde{\mathbf{x}}_{\sigma(t)}$ in Riemannian space were pre-generated, with three noisy structures for each reference structure in QM9 \cite{qm9}, ten for each in QM7-X \cite{qm7x}, and one for each in GEOM-QM9 \cite{GEOM}, for efficiency.

To effectively incorporate 3D molecular information, R-DM integrates SchNet layers with a 2D graph neural network architecture, primarily building on the design of TSDiff \cite{TSDiff} and GeoDiff \cite{GeoDiff}. All main results are obtained using the connectivity information, and the ablation without it is presented in the Supplementary Section~4. In addition, the ablation study comparing Euclidean and Riemannian components is provided in Supplementary Section~3 with a detailed discussion. Details on training hyperparameters related to E-DM and R-DM can be found in Supplementary Table~1.

\subsection{Measurement details} \label{sec:measurement_details}
We use four measures to evaluate molecular structure deviations: RMSD, D-MAE, geodesic distance, and absolute energy difference.
This section describes the details of each measurement.

RMSD is widely used to measure the average deviation between two structures, $\mathbf{x}$ (reference) and $\hat{\mathbf{x}}$ (predicted), by considering the root mean square of the differences in the atomic positions after optimal alignment using Kabsch algorithm \cite{kabsch1976solution}. The RMSD between $\mathbf{x}$ and $\hat{\mathbf{x}}$ is defined as:
\begin{equation}
    \text{RMSD}(\mathbf{x}, \hat{\mathbf{x}}) = \sqrt{\frac{1}{N_{\text{atom}}} \sum_{i=1}^{N_{\text{atom}}} \| \mathbf{x}_i - \hat{\mathbf{x}}_i \|_2^2 },
\end{equation}
where $\mathbf{x}_i$ and $\hat{\mathbf{x}}_i$ are the Cartesian position vectors of the $i$-th atom in $\mathbf{x}$ and $\hat{\mathbf{x}}$, respectively, and $N_{\text{atom}}$ is the number of atoms in each structure.

D-MAE computes the mean absolute deviation in interatomic distances using internal coordinates. The D-MAE between two different structures, $\mathbf{x}$ and $\hat{\mathbf{x}}$, is defined as
\begin{equation}
    \text{D-MAE}(\mathbf{x}, \hat{\mathbf{x}})
    = \frac{2}{N_{\text{atom}} (N_{\text{atom}}-1)}
    \sum_{i<j}^{N_{\text{atom}}} \abs{ \| \mathbf{x}_i - \mathbf{x}_j \|_2 - \| \hat{\mathbf{x}}_i - \hat{\mathbf{x}}_j \|_2 }.
\end{equation}

The geodesic distance measures the distance between two structures, $\mathbf{x}$ and $\hat{\mathbf{x}}$, on the manifold parameterized by a $q$-coordinate. The geodesic distance is computed using the geodesic interpolation method \cite{zhu2019geodesic}, in which the trajectory connecting $\mathbf{x}$ and $\hat{\mathbf{x}}$ is represented by a discrete series of $N$ images. This trajectory is optimized to become a minimal path, forming a geodesic line. For the optimized trajectory, the geodesic distance is then calculated as the sum of the lengths of the segments $l^{(n)}$, by
\begin{equation}
    L = \sum_{n=1}^{N-1} l^{(n)} =  \sum_{n=1}^{N-1} \| \mathbf{q}^{(n+1)} - \mathbf{q}^{(n)} \|_2,
\end{equation}
where $\mathbf{q}^{(n)}$ is the $n$-th interpolated image between $\mathbf{x}$ and $\hat{\mathbf{x}}$ represented on the $q$-coordinate.

The energy difference, $\abs{\Delta E}$, is calculated using single-point DFT calculations as follows
\begin{equation}
    \abs{\Delta E} = \abs{E(\mathbf{x}) - E(\hat{\mathbf{x}})},
\end{equation}
where $E(\cdot)$ is the total energy of the given structure. Detailed computational settings are provided in \cref{sec:computational_details}.

\subsection{Data}
The QM9 dataset \cite{qm9}, derived from the GDB-9 \cite{GDB17} chemical space, is a widely used benchmark in molecular machine learning, containing about 133k small organic molecules with up to nine heavy atoms (C, N, O, F). In the QM9 dataset, all molecular structures are optimized at the B3LYP/6-31G(2df,p) level, and the corresponding molecular properties are calculated at the same level. In this study, we used a random split of the QM9 dataset, assigning 100k molecules for training and 10k for testing.

In our evaluation on QM9, we utilize MMFF conformations corresponding to the reference DFT conformations as starting structures for R-DM and E-DM. The MMFF structures are sourced from the QM9$_\text{M}$ dataset \cite{qm9m}, where each structure is obtained by optimizing the original QM9 structures using the MMFF \cite{mmff} as implemented in RDKit \cite{landrum2006rdkit}. The QM9$_\text{M}$ dataset has been used in previous studies for property prediction, employing easy-to-obtain MMFF structures in place of computationally demanding DFT structures \cite{kim2024geotmi,lu2021dataset,pinheiro2022smiclr}.

The QM7-X dataset \cite{qm7x} contains 42,000 equilibrium molecular structures covering all molecular graphs in GDB-13 \cite{blum2009970} with up to 7 heavy atoms, including the elements H, C, N, O, S, and Cl. All equilibrium structures are optimized using third-order self-consistent charge density functional tight binding (DFTB3) \cite{gaus2011dftb3} with many-body dispersion (MBD) corrections \cite{tkatchenko2012accurate}. For each equilibrium structure, the dataset provides 100 non-equilibrium structures generated via normal-mode displacements of the equilibrium geometry. Following the data split used in MoreRed-JT \cite{morered}, we construct our test set by randomly selecting one non-equilibrium structure for each equilibrium structure, resulting in a total of 6,810 test samples.

The GEOM-QM9 dataset is a subset of the GEOM dataset \cite{GEOM} designed to capture the range of accessible 3D conformers for small, neutral molecules with up to nine heavy atoms. GEOM-QM9 provides individual 3D conformers for each molecule and an ensemble of conformations that reflect accessible molecular configurations. These conformers were generated with the CREST program \cite{crest}, which leverages semi-empirical tight-binding DFT (GFN2-xTB) \cite{gfn2} to produce reliable and diverse structures efficiently. Following previous works \cite{ConfGF, GeoDiff}, we assigned 40,000 molecules to the training set and 200 molecules to the test set. For each training molecule, we selected five representative conformations, resulting in a total of 200,000 conformations used in training.

\subsection{Computational details} \label{sec:computational_details}
We employed quantum chemical calculations to evaluate the energy accuracy. For the QM9 dataset, we used Gaussian 09, \revision{Revision D.01} \cite{Gaussian09} to perform single-point and structure optimization calculations with the same computational options as in the original work \cite{qm9}. Specifically, B3LYP/6-31G(2df,p) DFT was employed with the following options: opt(calcfc, maxstep=5, maxcycles=1000), integral(grid=ultrafine), and scf(maxcycle=200, verytight). Each calculation was conducted with four cores on an Intel(R) Xeon(R) CPU E5-2667 v4 @ 3.20GHz node using the nprocshared=4 option. For the QM7-X dataset, following the computational protocol established in MoreRed-JT \cite{morered}, we calculated molecular energies using PBE0+MBD with the def2-TZVP basis set \cite{def2_tzvp}, implemented in PySCF \cite{pyscf}. For the ensemble property evaluation, we used the PSI4 toolkit \cite{smith2020psi4} to calculate the energy and the HOMO-LUMO gap for each conformer, applying the default computational settings consistent with previous studies \cite{ConfGF,GeoDiff}.

For generation tasks, we used three generative methods: ETKDG \cite{ETKDG}, ConfGF \cite{ConfGF}, and GeoDiff \cite{GeoDiff}. The ETKDG \cite{ETKDG} and MMFF \cite{mmff} algorithms, implemented in RDKit \cite{landrum2006rdkit} (version 2020.09.1), were used. We also employed the trained ConfGF and GeoDiff models, obtained from their respective repositories at \url{https://github.com/DeepGraphLearning/ConfGF} and \url{https://github.com/MinkaiXu/GeoDiff/tree/pretrain}.

\section{Data availability}
The QM9 dataset \cite{qm9} is accessible at \url{https://doi.org/10.1038/s41597-022-01288-4}. The QM9$_\text{M}$ dataset \cite{qm9m}, which includes MMFF structures corresponding to QM9 structures, is available at \url{https://doi.org/10.1021/acs.jctc.9b00001}. We used the GEOM \cite{GEOM} dataset, preprocessed and available at \url{https://github.com/MinkaiXu/GeoDiff/tree/pretrain}. \revision{Source data for Figures 1–3 are available with this manuscript.}

\section{Code availability}
An implementation of the proposed model, R-DM, is publicly available under the MIT License at GitHub (\url{https://github.com/seonghann/neural_opt/tree/refactoring}) and has been archived on Zenodo for long-term accessibility \cite{rdsm_github}.

\section{Acknowledgments}
This work was supported by the Korea Environmental Industry and Technology Institute (Grant No. RS202300219144, to J. Woo, S. Kim, J. H. Kim, and W. Y. Kim). It was also supported by the InnoCORE program of the Ministry of Science and ICT (Grant No. N10250153, to S. Kim), and the NRF funded by the Ministry of Science and ICT (Grant No. NRF-2022M3J6A1063021, to W. Y. Kim).

\section{Author information}
These authors contributed equally: Jeheon Woo, Seonghwan Kim.

\subsection{Contributions}
J. Woo and S. Kim contributed equally to this work. They designed the methodology and the experiments. J. H. Kim performed the conformer generation experiments. All authors wrote the manuscript together, and W. Y. Kim supervised the project.

\subsection{Corresponding author}
Correspondence to Woo Youn Kim.

\section{Ethics declarations}
\subsection{Competing interests}
The authors declare no competing interests.


\newpage
\section{Tables}
\begin{table}[htbp]
\centering
\caption{
\textbf{Comparison of RMSD, D-MAE, and $\abs{\Delta E}$ (kcal/mol) for different models.}
This table shows the root mean square deviation (RMSD), the mean absolute error of the interatomic distances (D-MAE) and the energy difference ($\abs{\Delta E}$), using the DFT structures as references for those optimized by MMFF, E-DM, and R-DM. The evaluation was performed on the QM9 test set, and the energy values were obtained from single-point DFT calculations. Bold symbol indicate the best performance for each metric.
} \label{tab:tab1}
\resizebox{0.8\textwidth}{!}{
\begin{tabular}{lllllll}
\hline
 & \multicolumn{2}{c}{RMSD (\AA)} & \multicolumn{2}{c}{D-MAE (\AA)} & \multicolumn{2}{c}{$\abs{\Delta E}$ (kcal/mol)} \\
 & Mean & Median & Mean & Median & Mean & Median \\
\hline
MMFF & 0.200 & 0.137 & 0.0717 & 0.0571 & 7.386 & 4.215 \\
E-DM & 0.166 & 0.091 & 0.0454 & 0.0280 & 1.560 & 0.816 \\
R-DM & \textbf{0.104} & \textbf{0.031} & \textbf{0.0256} & \textbf{0.0095} & \textbf{0.806} & \textbf{0.177} \\
\hline
\end{tabular}
}
\end{table}

\begin{table}[htbp]
\centering
\caption{
\textbf{Comparison of the DFT optimization cost for structures obtained by MMFF and R-DM.}
DFT optimizations were conducted on 1,000 randomly selected molecules from the QM9 test set using initial structures obtained by MMFF and R-DM. The number of force calls, total number of cycles and time of self-consistent field (SCF) are compared. Statistics were computed for the 964 cases where the initial structures from both MMFF and R-DM converged to the same final structures.
}
\label{tab:post_opt}
\begin{tabular}{lrrrrrr}
\hline
& \multicolumn{2}{c}{\# of force calls} & \multicolumn{2}{c}{\# of SCF cycles} & \multicolumn{2}{c}{SCF time (s)} \\
& MMFF & R-DM & MMFF & R-DM & MMFF & R-DM \\
\hline
Mean  & 9.1 & 4.4 & 95.4 & 45.6 & 1200.0 & 573.0 \\
Q1  & 7 & 3 & 67 & 29 & 850.2 & 356.4 \\
Q2 (Median) & 8 & 3 & 87 & 35 & 1110.0 & 472.5 \\
Q3  & 11 & 5 & 114 & 49 & 1431.5 & 643.5 \\
\hline
\end{tabular}
\end{table}

\begin{table}[htbp]
\centering
\caption{
\textbf{Comparison of optimization performance on QM7-X dataset.}
Median values of RMSD, D-MAE, and absolute energy difference ($\abs{\Delta E}$) are reported for optimizations starting from non-equilibrium initial structures (Non-eq). R-DM$^\ddagger$ denotes R-DM using MoreRed-JT results as initial structures.
All energies are calculated using PBE0+MBD level of theory.
}
\label{tab:qm7x_results}
\begin{tabular}{llll}
\hline
 & RMSD (\AA) & D-MAE (\AA) & $\abs{\Delta E}$ (kcal/mol) \\
\hline
Non-eq & 0.341 & 0.1564 & 51.814 \\
E-DM & 0.122 & 0.0319 & 0.610 \\
R-DM & 0.046 & 0.0121 & 0.277 \\
\hline
MoreRed-JT & 0.050 & 0.0165 & 0.269 \\
R-DM$^\ddagger$ & 0.034 & 0.0102 & 0.150 \\
\hline
\end{tabular}
\end{table}

\clearpage
\section{Figure Legends/Captions}

\begin{figure}[htbp] 
\centering
\includegraphics[width=1.0\textwidth]{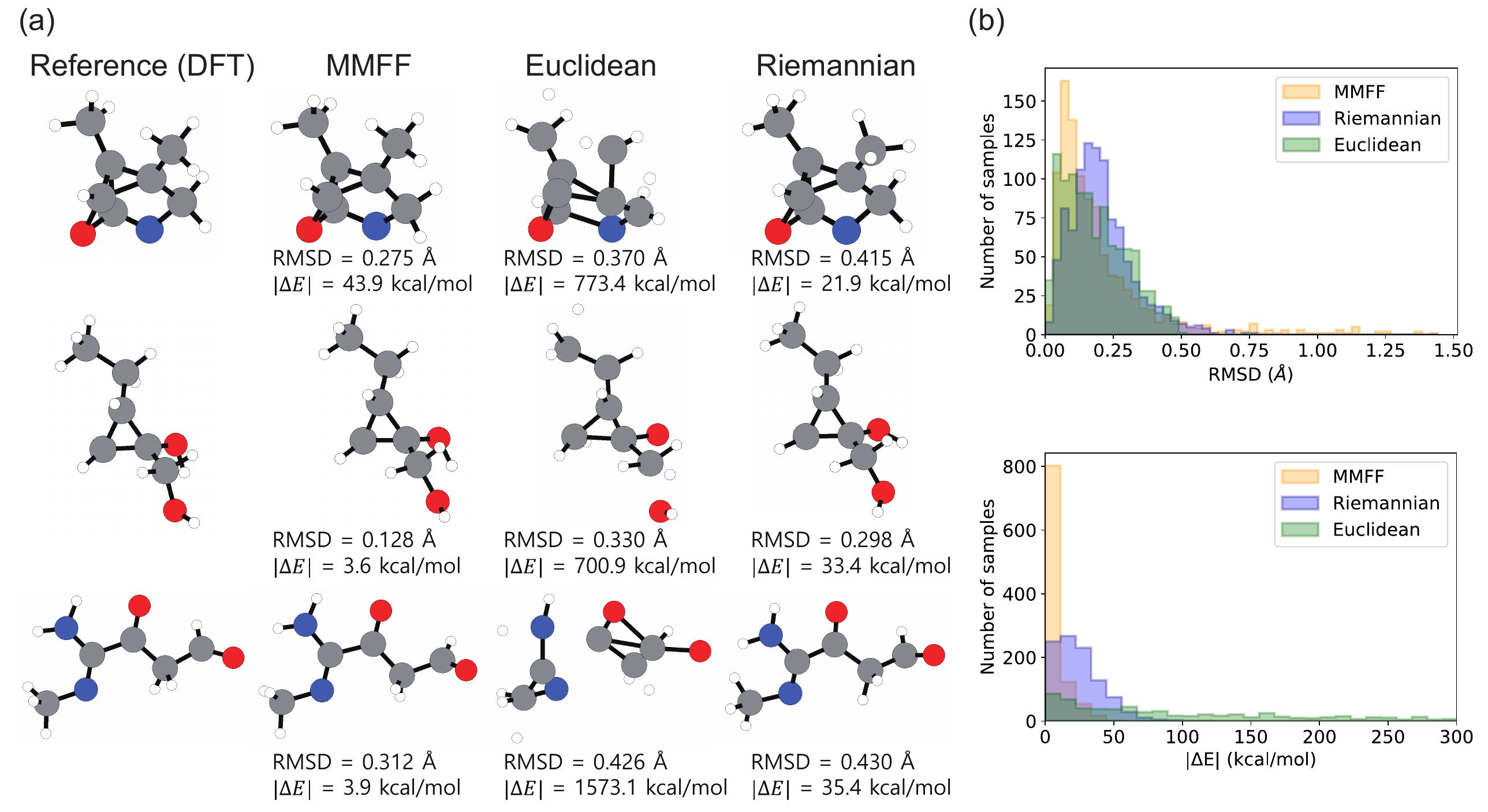}
\caption{
\textbf{Comparison of molecular structures sampled from noisy distributions in Euclidean and Riemannian spaces.}
\textbf{a} Visualization of three example molecular structures obtained from each method: from left to right, DFT, MMFF, Euclidean noise-sampling, and Riemannian noise-sampling.
For each structure, the root mean square deviation (RMSD) and $\abs{\Delta E}$ values are shown with respect to the reference DFT structure.
\textbf{b} The distributions of RMSD (top) and $\abs{\Delta E}$ (bottom) for molecular structures obtained by MMFF (orange), Riemannian noise-sampling (blue), and Euclidean noise-sampling (green), with the DFT structures as reference.
Each distribution is plotted with 1,000 molecules randomly selected from each method; however, 23 samples were excluded from the Euclidean noise-sampled structures due to failure in self-consistent field (SCF) convergences. \revision{Atom colors follow the standard convention: C, gray; H, white; O, red; N, blue.}
}\label{fig:noisy_distribution}
\end{figure}

\begin{figure}[htbp] 
\centering
\includegraphics[width=0.5\textwidth]{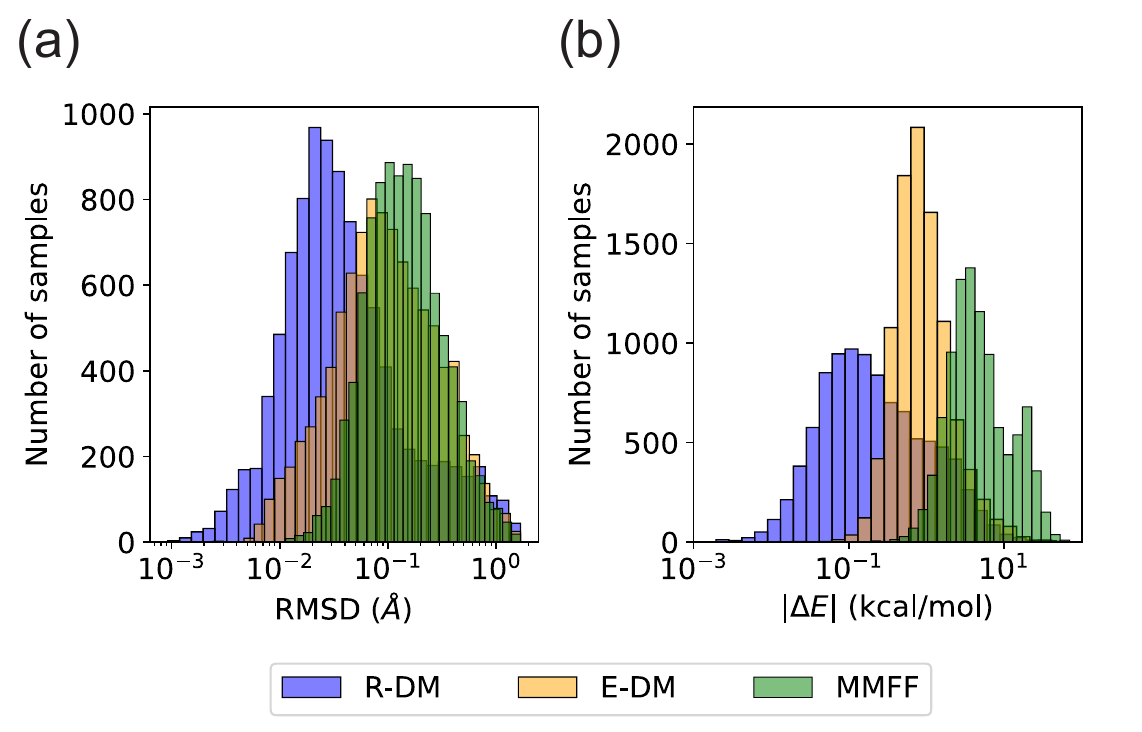}
\caption{
\textbf{Comparative analysis of molecular structures predicted by R-DM, E-DM, and MMFF.}
Histograms compare the distributions of (a) root mean square deviation (RMSD) and (b) absolute energy difference ($\abs{\Delta E}$) between molecular structures predicted by R-DM (blue), E-DM (yellow), and MMFF (green). The sampling was performed on the QM9 test set, and the energy values were obtained from single-point DFT calculations.
}\label{figure3}
\end{figure}

\clearpage
\includepdf[pages=1-20]{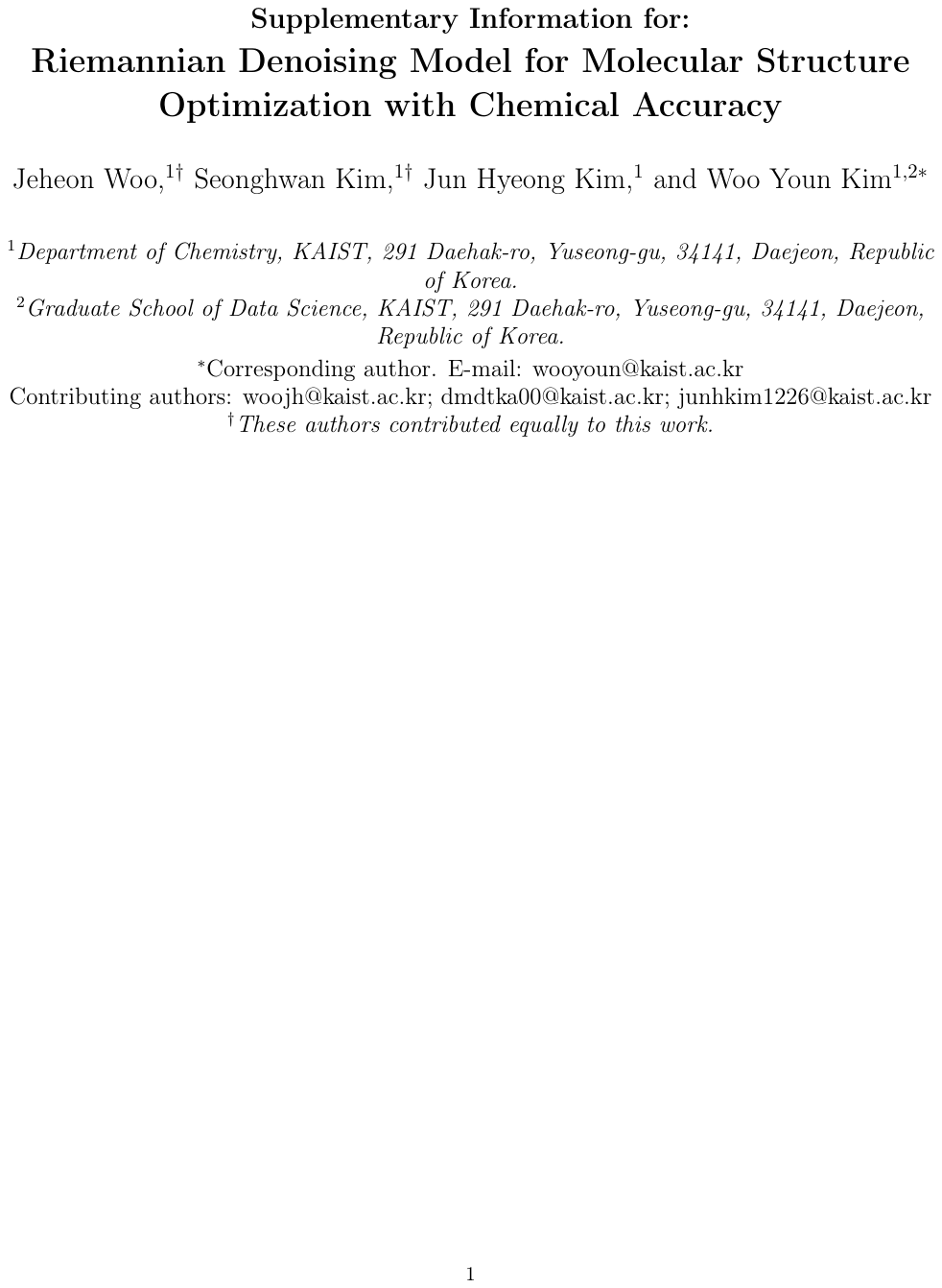}

\end{document}